\documentclass[review, compsoc]{IEEEtran}
\usepackage[comma,numbers,square,sort&compress]{natbib}
\usepackage{amsfonts,amssymb,amsthm}
\usepackage{amsmath}
\usepackage{epsfig}
\usepackage{float}
\usepackage{graphicx}
\usepackage{color,hyperref}
\usepackage{multirow}
\usepackage{tipa}
\usepackage{picins}
\usepackage{color}
\usepackage{booktabs}
\usepackage[table]{xcolor}
\usepackage{threeparttable}

\begin{document}

\title{UAV Obstacle Avoidance by Human-in-the-Loop Reinforcement in Arbitrary 3D Environment}

\author{Xuyang Li, Jianwu Fang, Kai Du, Kuizhi Mei, and Jianru Xue 
\thanks{X. Li, J. Fang, and K. Du are with the Chang'an University, Xi'an, China
        {(fangjianwu@chd.edu.cn)}.}%
\thanks{K. Mei and J. Xue are with the Institute of Artificial Intelligence and Robotics, Xi'an Jiaotong University, Xi'an, China
        {(jrxue@mail.xjtu.edu.cn)}.}%
}

\IEEEtitleabstractindextext{%
\begin{abstract}
This paper focuses on the continuous control of the unmanned aerial vehicle (UAV) based on a deep reinforcement learning method for a large-scale 3D complex environment. The purpose is to make the UAV reach any target point from a certain starting point, and the flying height and speed are variable during navigation. In this work, we propose a deep reinforcement learning (DRL)-based method combined with human-in-the-loop, which allows the UAV to avoid obstacles automatically during flying. We design multiple reward functions based on the relevant domain knowledge to guide UAV navigation. The role of human-in-the-loop is to dynamically change the reward function of the UAV in different situations to suit the obstacle avoidance of the UAV better. We verify the success rate and average step size on urban, rural, and forest scenarios, and the experimental results show that the proposed method can reduce the training convergence time and improve the efficiency and accuracy of navigation tasks. The code is available on the website \url{https://github.com/Monnalo/UAV_navigation}. 
\end{abstract}

\begin{IEEEkeywords}
Deep reinforcement learning, obstacle avoidance, UAV, human-in-the-loop, POMDP
\end{IEEEkeywords}}
\markboth{}%
{}
\maketitle

\section{Introduction}

Unmanned aerial vehicles (UAVs) have developed rapidly in the past few years and are important for many applications, such as smart cities \cite{1}, aerial photography \cite{2}, and emergency rescue \cite{3}. The development of UAVs' autonomy has become the key to adapting to various complex environments. The improvement of autonomy cannot only save manpower but also improve the efficiency of specific tasks. It is a necessary and challenging problem for the path planning of UAVs when they perform different missions, especially in a complex 3D environment.

Traditional methods in UAV navigation are generally non-learning methods that commonly take various sensors to perceive the obstacles around UAVs, and then use the path planning to navigate the UAVs \cite{4,5,6}. Another navigation solution for UAVs is building the environment map through simultaneous localization and mapping (SLAM) \cite{8,9}. These methods require robust path planning when navigating, and are difficult to be implemented in highly complex 3D scenes.

The rise of reinforcement learning recently provides a new way for UAV navigation. For example, coverage path planning problem \cite{10,11,12} is concerned with rasterizing the perceived environment map, and the motion of the UAV is simplified as an optimal grid moving problem. At this time, a reinforcement learning algorithm for discrete control problems is applied. Similar to this idea, when studying the obstacle avoidance problem of UAVs, some works characterize the action space of UAVs as a set of different motion directions. For example, when the UAV moves in each time step, a fixed distance length is selected from several directions such as move front, move back, turn left, and turn right \cite{13,14}. Some other methods also set several different moving speeds when controlling the movement of UAVs in different directions \cite{15,16}. However, in these methods, the movement of the UAV is limited to a plane with a constant height, and the control strategy is similar to one of the vehicles on the road. The UAVs do not play their flexibility of flying at different heights.
\begin{figure}[t]
\centering
\includegraphics[width=70mm,height=60mm]{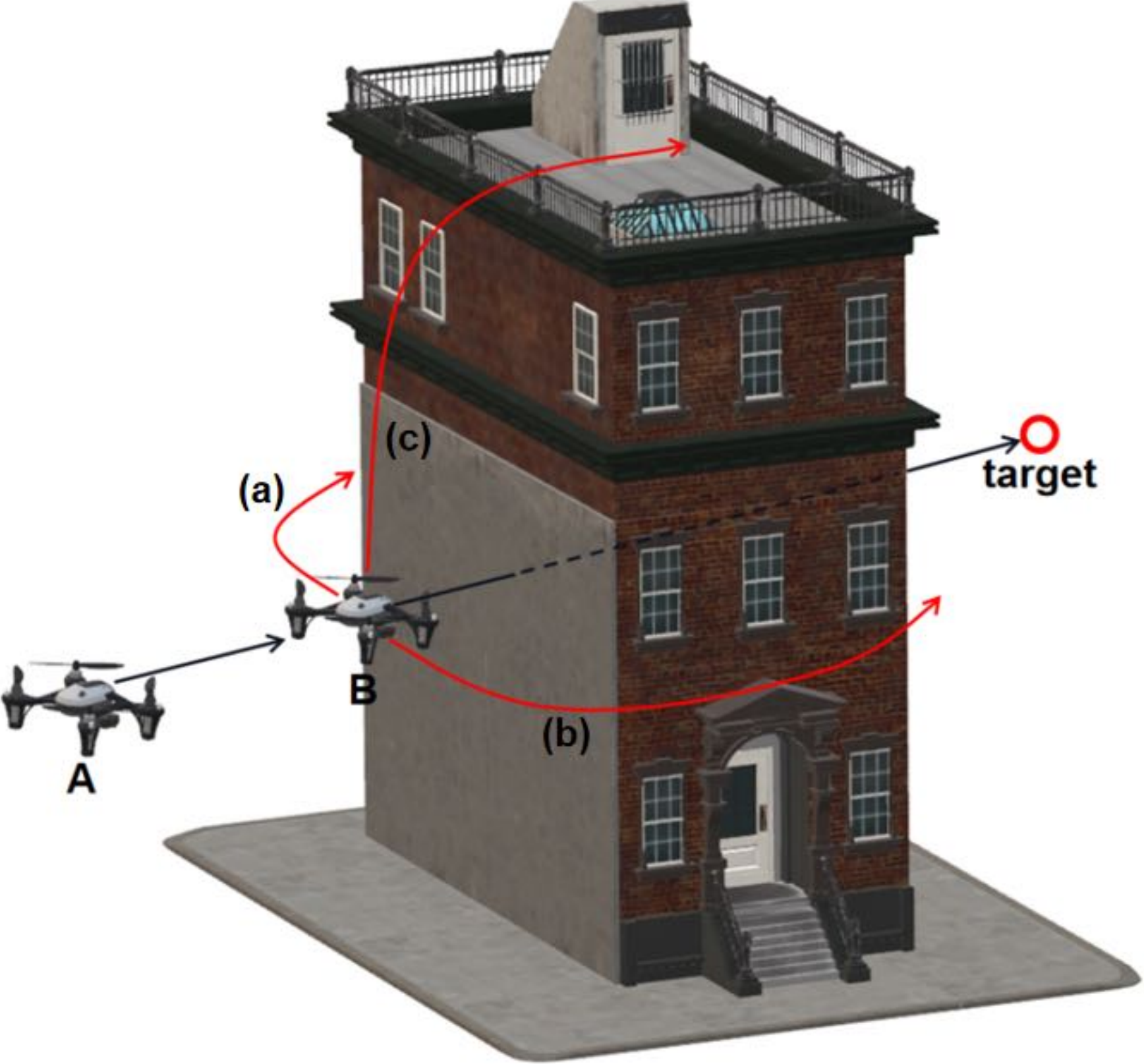}
\caption{\small{The obstacle avoidance of a UAV in a 3D environment. There are different choices of obstacle avoidance for the UAV. When this UAV is at the starting point $A$ (no obstacle), it will fly forward normally. When the UAV arrives at point $B$ (there is a building ahead), the UAV needs to avoid the building when navigating. For two-dimensional (2D) navigation tasks, UAV can generally avoid the building from the direction of ($a$) or ($b$), while for 3D navigation tasks, the direction of ($c$) is also an alternative, which will increase the action space of the UAV.}}
\label{fig1}
\end{figure}

\begin{figure*}
\centering
\includegraphics[width=175mm,height=70mm]{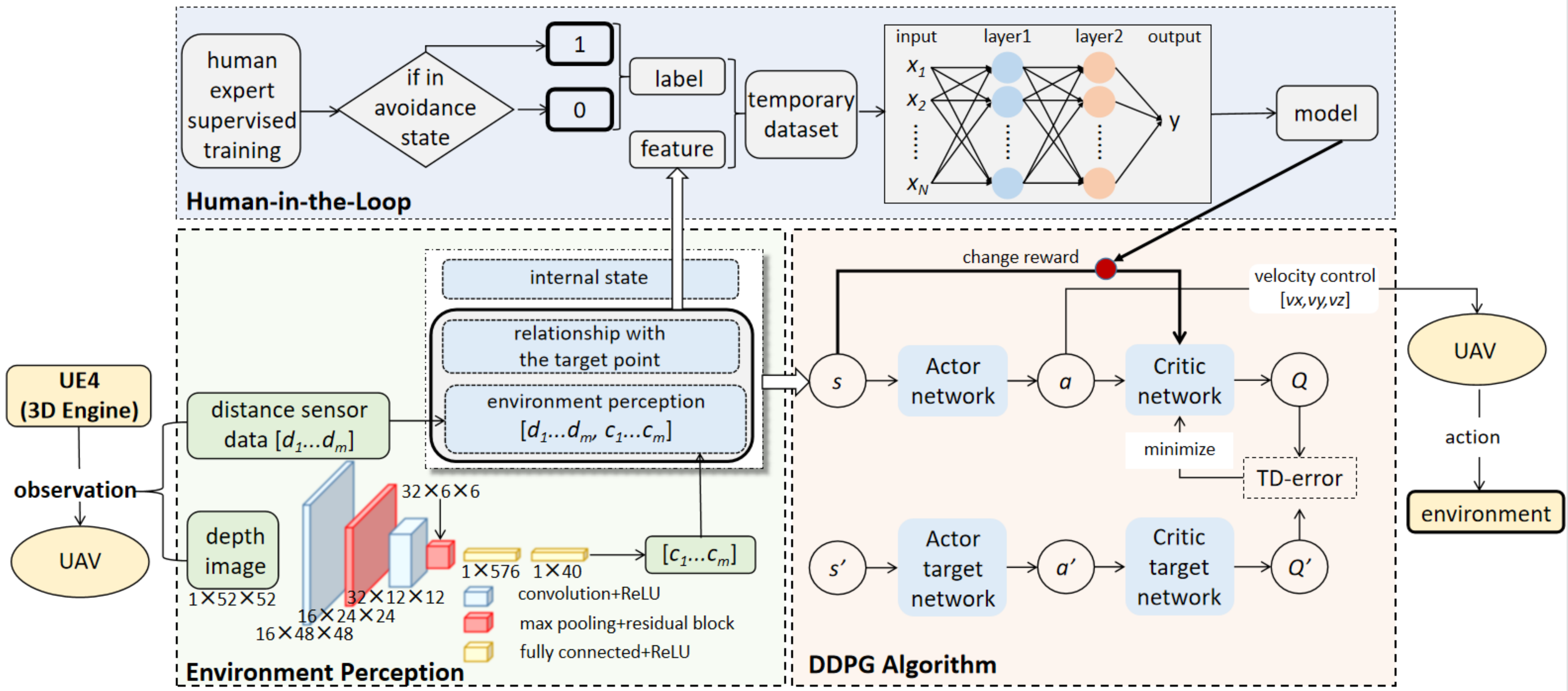}
\caption{\small{The flowchart of the proposed method. Firstly, the UAV senses environment information through a depth camera and $m$ distance sensors with different perception directions. The distance sensors will return a set of distance values $[d_1...d_m]$. The depth image returned by the depth camera is encoded by a pre-defined convolution neural network (CNN), and the $n$-dimensional feature embedding is $\emph{\bf{c}}=[c_1...c_n]$. The above information constitutes the internal characteristics of the UAV, and the relationship between the UAV and the target point together, which is further treated as the state $s$ in the deep reinforcement learning framework (DRL). We design the deep deterministic policy gradient (DDPG) algorithm and twin delayed deep deterministic policy gradient algorithm (TD3) to control the three-way speed of the UAV. To be clear, we show the DDPG flowchart in this figure. During reinforcement learning training, the UAV's perception of the environment $[d_1...d_m,c_1...c_n]$ and its relationship feature $(d_{dist},y_{yaw})$ with the target point form a feature vector $[d_1...d_m,c_1...c_n,d_{dist},y_{yaw}]$. Specially, we introduce the human expert to assist the avoidance state recognition in DDPG and TD3 and update the avoidance state label of the extracted feature vectors, which can help the UAV to avoid obstacles more efficiently. }}
\label{fig2}
\end{figure*}

In addition, some methods consider the problem of height change when controlling UAVs. For example, in solving the obstacle avoidance problem of UAVs in tunnel environments, one idea is to give the UAV a fixed forward speed and control the UAV to move up, down, left, or right to avoid obstacles \cite{17}. There are also some ways to extend the UAV's action space to 3D space \cite{18,19}. For example, when the UAV is moving, it may move up, down, or obliquely. However, the number of moving directions in a 3D environment is difficult to be designed. If the types of moving direction in the action space are too few, the motion of the UAV is difficult to meet the actual path. If the action space contains too many types of moving directions, the efficiency of the algorithm will be degraded. Therefore, regarding the control of the UAV as a continuous control problem (given the instruction is a continuously variable speed, acceleration or heading, \emph{etc}.), we will make the UAV motion space more realistic and guarantee efficiency. For example, two works based on DDPG \cite{20} and recurrent deterministic policy gradient (RDPG) \cite{21} set a constant speed for the UAV when studying the 3D obstacle avoidance problem, and then output a continuous variable heading through the reinforcement learning framework. However, all of the above works are non-autonomous in speed control. In this case, when there are other fast-moving objects in the scene, the constant-speed flying UAV may face a series of problems.

To sum up, in order to fit the real 3D environment, we build a high-fidelity virtual environment, in which the obstacle avoidance problem of UAVs is studied. Based on the DDPG \cite{22} and TD3 \cite{23},  we re-design the reward function and action space for continuous control of UAVs. In order to make the UAV explore an effective navigation policy quickly in large action space, we introduce an \emph{avoidance state} \footnote{When the UAV needs to make an obstacle avoidance, we call the state of the UAV at this time as an \emph{avoidance state}.} for the UAV, and dynamically change the reward function in this state to guide the UAV movement better. However, due to the complexity of the 3D environment and the limited sensing ability of the UAV, it is difficult to additionally recognize the state while exploring the navigation policy. Therefore, we introduce the human role to contributing expert knowledge to assist UAVs in identifying the \emph{avoidance state}. We call the DDPG and TD3 combined with the human-in-the-loop as DDPG-H and TD3-H (see Section 3.2 for detail). To be clear, Fig. \ref{fig2} presents the flowchart of this work.

The main \textbf{contributions} of this paper are as follows:
\begin{enumerate}
\item DDPG-H and TD3-H are proposed to continuously control the UAV in 3D environments with different heights and speeds by introducing the human expert knowledge in recognition of the \emph{avoidance state}.
\item Through the human-in-the-loop mechanism, we can dynamically change the reward function, so as to better guide the UAV navigation. The superiority of the proposed method is verified in the experiments on urban, rural, and forest scenes.
\end{enumerate}

\section{Problem Formulation}
\subsection{Problem Description}

We use the four-rotor UAV provided by the virtual simulation platform Airsim \cite{24} to study in the virtual 3D environment built by UE4 \cite{25}. We ignore the momentum when the UAV is flying, \emph{i.e.}, the turning of the UAV can work immediately. In addition, we assume that the UAV can obtain its own position and speed information in real-time.

The goal of the navigation of UAV starts from a starting point $\emph{\bf{s}} = [x_s, y_s, z_s] $ in a 3D environment (about 1.4 square kilometers) and flies to a random target point $\emph{\bf{o}} = [x_o, y_o, z_o] $. In this process, assume that UAV does not cause a collision with the obstacle and will not cross the environment boundary. In this problem, UAVs cannot get global and local information on the map in real-time. The environment perception depends on the onboard camera or other distance sensors. This kind of observation approach has perception uncertainty. Therefore, the obstacle avoidance problem is actually a kind of partially observable Markov decision process (POMDP) \cite{26}. Because the weight, volume, and cost of different sensors vary greatly in reality, it is necessary to check the influence of the observation by different kinds of sensors. Therefore, based on the same baseline, we evaluate different sensor configurations of the UAV in the experiments. In addition, we hope to study the performance improvement of the algorithm by changing the reward function, so we test the performance of the initial algorithm and the algorithm combined with human-in-the-loop.

\subsection{Observation Space and Action Space}

In order to fulfill the navigation task, the UAV should be able to obtain the following data: the internal state, the 3D environment information, and the relationship with the intended target point. Due to the complexity of the 3D environment, we use the UAV orientation $\emph{\bf{p}}=[p_x, p_y, p_z]$ and three-way speed $\emph{\bf{v}}=[v_x, v_y, v_z]$ to describe the internal state of the UAV, denoted as $\phi = [\emph{\bf{p}}, \emph{\bf{v}}] $. For the perception of the 3D environment, we use the depth image and the distance to obstacles returned by the UAV depth camera and distance range sensors (12 distance sensors with different directions are used) to characterize the observation space. The data obtained by $m$ distance sensors is denoted as $\emph{\bf{d}}=[d_1...d_m]$, and the depth image is encoded by a pre-trained CNN model and the feature embedding is denoted as a $n$-dimensional vector $\emph{\bf{c}}=[c_1...c_n]$. CNN model here is illustrated in Fig. \ref{fig2} with some interleaved layers of two \emph{convolution+Relu} ($16\times48\times48$ and $32\times12\times12$), one \emph{maxpooling + residual block} ($downsampling\downarrow2$), and two \emph{fully connected layer+ReLU}. For the relationship with the intended target point, the distance and heading angle between the UAV and the target point are used and denoted as $\zeta = [d_{dist}, y_{yaw}]$.

In order to make the height and the speed of the UAV in the 3D environment changeable, we directly control the 3D-axis velocity of the UAV and the orientation of the UAV changes with the direction of the total velocity. We learn the 3D-axis acceleration of the UAV through the deep reinforcement learning framework, \emph{i.e.}, performing the action space $\emph{\bf{a}}= [a_x, a_y, a_z] $. At each time step, the current flight speed $\emph{\bf{v}}_t= [v_{x_t}, v_{y_t}, v_{z_t}]$ is obtained, and the output acceleration $\emph{\bf{a}}_t = [a_{x_t}, a_{y_t}, a_{z_t}]$ is added to obtain the speed in next time $\emph{\bf{v}}_{t+1} = [v_{x_{t+1}}, v_{y_{t+1}}, v_{z_{t+1}}]$. We send a speed control command to the UAV at an interval of fewer than 0.5 seconds. Because of the small interval between every two steps, we can continuously control the UAV smoothly.

\subsection{Reward Design}

The reward is a formal and numerical representation for approximating the intended target by the UAV. In reinforcement learning, the simplest way is to define the intended goal as a sparse reward \cite{27,28}. That is to say that the UAV will be rewarded when it successfully reaches the target point. However, in a complex 3D environment with a large state-action space, the probability of reaching their intended goals through random policy in the early stage is extremely small, and sparse rewards cannot indicate the exploration direction of UAVs. In this case, reinforcement learning needs a long time to converge or even is unable to converge. Therefore, another approach is to add auxiliary rewards to the original sparse rewards. Generally, the auxiliary rewards are with the non-sparse property. In this paper, we set some sparse rewards and non-sparse rewards based on the knowledge of UAV navigation. 

\textbf{Sparse rewards}: Sparse rewards include \emph{arrival reward}, \emph{collision penalty}, and \emph{out-of-range penalty}. The arrival reward $r_{arri}$ is defined as a positive constant $\sigma$.
 Besides, during navigation, we want the UAV to avoid the collision or leave the working area as much as possible. Hence, when these behaviors occur, UAV will be punished, and we design the collision penalty $r_{coll}$ and out-of-range penalty $r_{out}$ also as a positive constant $\eta$. 

\textbf{Non-sparse rewards}: Since the UAV's navigation to the target point may take a long process, the above non-sparse rewards alone cannot achieve the expected purpose. Therefore, we set some non-sparse rewards to guide the UAV better. First of all, we hope that the moving direction of the UAV is without large bias, so when the heading angle deviation of the UAV is large, we will give the \emph{yaw penalty}. 

\emph{yaw penalty}: The yaw penalty is designed as:
\begin{equation}
  \label{eq3}
  r_{yaw} =
  \begin{cases}
  0, &{|yaw| < \psi}\\
  \frac{- \xi * yaw}{180}, &{|yaw| \geq \psi}
  \end{cases},
\end{equation}
where $\xi$ is a penalty coefficient to characterize the yaw penalty. $\psi$ is an angle deviation threshold, and we encourage the angle deviation of UAVs to remain within $\psi$. 

\emph{wander penalty}: In order to encourage the UAV to find a better path to reach the target point as soon as possible, we design a \emph{wander penalty}. As long as the UAV does not reach the target point, it will continue to be punished. The \emph{wander penalty} is designed as follows:
\begin{equation}
  \label{eq4}
  r_{wand} = \frac{- \delta * \lambda}{d_{dist}},
\end{equation}
\begin{equation}
  \label{eq5}
  d_{dist} =
  \begin{cases}
  \lambda, &{d_{dist} < \lambda}\\
  d_{dist}, &{d_{dist} \geq \lambda}
  \end{cases} 
	,
\end{equation}
where $\delta$ is a coefficient to characterize the wander penalty. 

\emph{height penalties}: In addition, because the flight distance is different during each navigation step, \emph{wander penalty} should vary with flight distance. Assume $d_{dist}$ is the distance between the starting point and the target point. This may result in a large penalty if the navigation distance is too short, so a distance threshold $\lambda$ is added to limit the penalty. Likewise, we do not want UAVs to fly too high, which could lead the UAV to avoid obstacles from the top no matter what obstacles it encountered. Therefore, when the height difference between the UAV and the target point is large, the \emph{height penalty} is given. The definition of \emph{height penalty} is as:
\begin{equation}
  \label{eq6}
  r_{z_{dist}} =
  \begin{cases}
  0, &{|z_{dist}| < \beta_1}\\
  - \alpha_1, &{\beta_1 \leq |z_{dist}| < \beta_2}\\
  - \alpha_2, &{|z_{dist}| \geq \beta_2}
  \end{cases},
\end{equation}
where $z_{dist}$ represents the height difference between the current location of the UAV and the target point. $\alpha$ and $\beta$ are used to give different penalties for different height differences. In summary, the final non-sparse reward can be expressed as follows:
\begin{equation}
  \label{eq7}
  r_{final} = r_{wand} + r_{yaw} + r_{zdist}.
\end{equation}

\section{DDPG-H And TD3-H For Navigation}

In this section, we first describe how the deep reinforcement learning algorithm is used in the navigation task, and then we describe the method of dynamically changing the reward function if needed through the human-in-the-loop mechanism.
\begin{figure}[t]
\centering
\includegraphics[width=80mm,height=60mm]{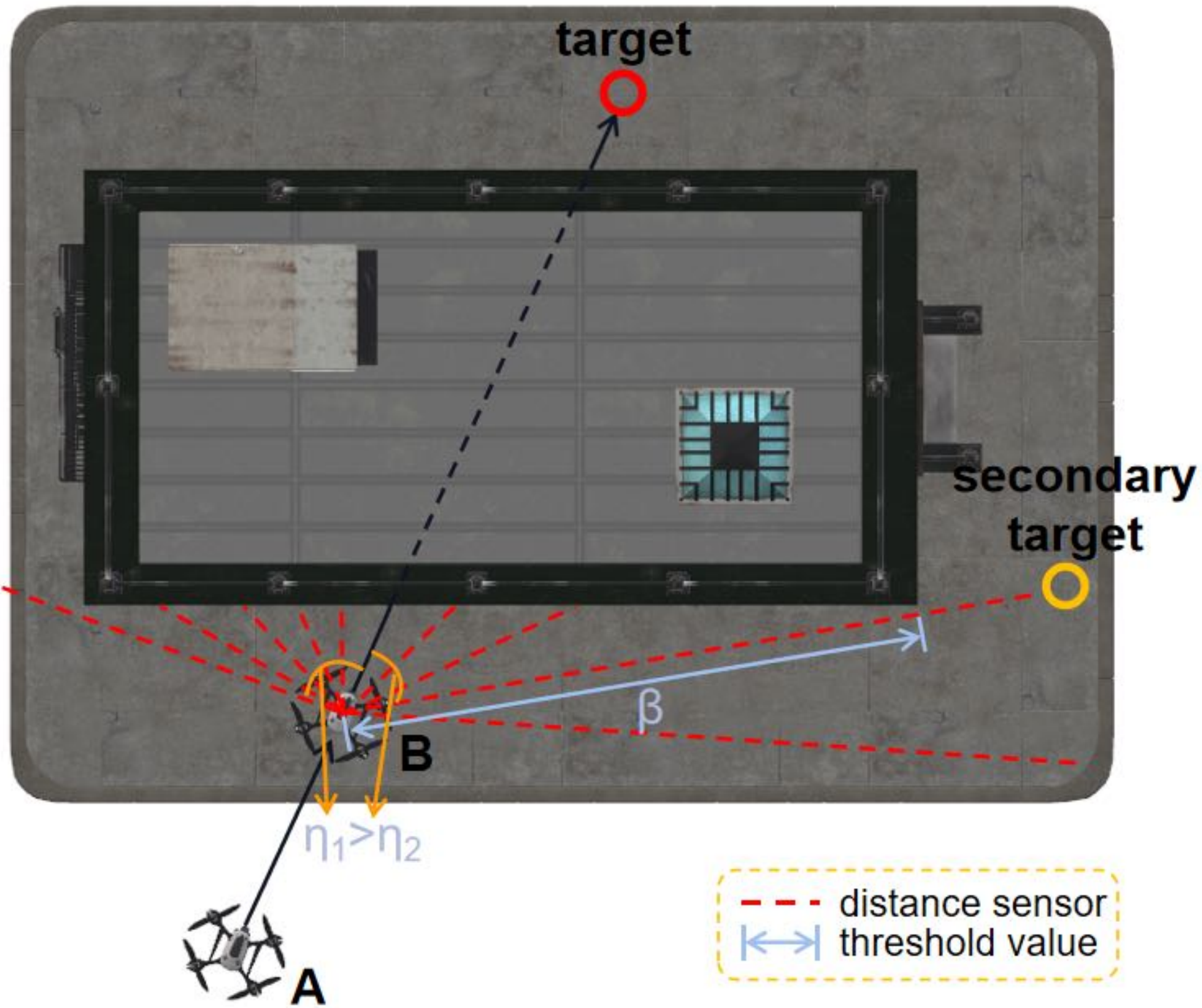}
\caption{\small{UAV's secondary target point selection in navigation. When the UAV arrives at point B, the UAV needs to make obstacle avoidance. We place $m$ distance sensors $[d_1,... d_m]$ around the UAV sensing plane and set a threshold value for collision avoidance. Among the sensors whose distance is larger than the pre-defined threshold $\beta$ (25 meters in the urban scene, and 10 meters in the other scenes), the distance sensor $d_m$ with the smallest angle $\eta$ to the direction of the target point is selected, and a certain distance is extended along the direction of the sensor to obtain the secondary target point with the consistent height to the target point.}}
\label{fig3}
\end{figure}

\subsection{DDPG and TD3}
DDPG is based on the Actor-Critic framework \cite{29}. Because of the deterministic policy, they can reduce the number of samples required in the high-dimensional action space. As shown in Fig. \ref{fig2}, DDPG can be divided into the Actor policy network and Critic value network, and each network is subdivided into the estimation network and target network. The role of the Actor-network is to output the deterministic actions by adopting deterministic policies. The estimation network of the Actor is used to output the real-time actions, and the target network of the Actor is used to update the Critic value network. A Critic network is used to fit the value function, which also has an estimation network and a target network. Critic's target network has two input parameters, which are the observation of the current state $s$ and the action $a$ of the output of the Actor's target network. The input of Critic's target network is the action $a$ of the output of the current Actor's estimated network.

TD3 is improved on the basis of DDPG, and three important optimizations are made to DDPG. The first is that TD3 uses two sets of Critic networks to estimate the Q value and use the smaller Q value as the update target. This is done to solve the situation where Q is overvalued. The second improvement is to stabilize the training process by setting delayed updates. A final improvement is to add a small amount of random noise to the target action.

\subsection{Reward Recombination}
In complex reinforcement learning tasks, it is often difficult to avoid the interference of non-sparse rewards in the pursuit of the maximization of total reward, which results in the destruction of optimal policies. With the reward functions aforementioned in this work, when an obstacle appears in front of the UAV, we want the UAV to avoid the obstacle. However, the UAV will be subject to a yaw penalty when conducting the avoidance. Especially, for some large obstacles, the UAV will receive more punishment when performing the avoidance. Therefore, we plan to use several obstacle avoidance tasks performed by the UAV during autonomous navigation as sub-tasks and modify the initially designed reward function when performing sub-tasks.

For ease of description, we transfer the view in Fig. \ref{fig1} into the bird's eye view (BEV), as shown in Fig. \ref{fig3}. When the UAV is in an avoidance state, a secondary target point is set for the UAV via the UAV's distance sensors. At this time, we remove the yaw penalty. In addition, a non-sparse reward function is added to encourage the UAV to move to the second target point. The reward function is defined as:
\begin{equation}
  r_{sec} =
  \begin{cases}
  \mu_2, &{|s_{yaw}| < \xi_1}\\
  \mu_1, &{\xi_1 \leq |s_{yaw}| < \xi_2}\\
  0, &{|s_{yaw}| \geq \xi_2}
  \end{cases}
    \label{eq8}
\end{equation}
Therefore, in the avoidance state, the sum of non-sparse rewards of Eq. \ref{eq7} for UAV is changed to:
\begin{equation}
  \label{eq9}
  r_{final} = r_{sec} + r_{zdist} + r_{wand}
\end{equation}

\begin{figure}[!t]
\centering
\includegraphics[width=80mm,height=35mm]{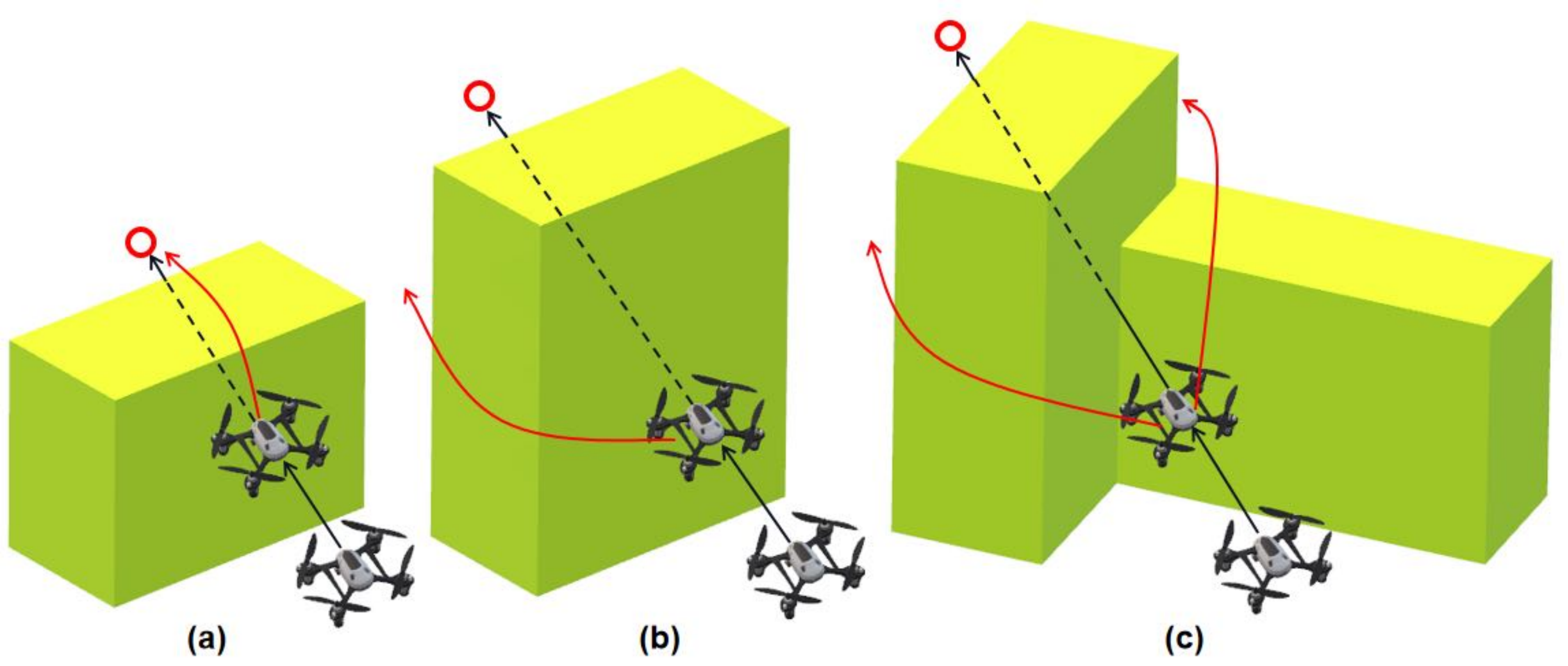}
\caption{\small{3D obstacles with different heights. When the obstacle is low shown in (a), we want the UAV to cross from top space. When the obstacle is high shown in (b), we want the UAV to cross from both sides. For irregular obstacles, e.g., (c), it is a good choice for the UAV to avoid obstacles either in the left or the top right directions.}}
\label{fig4}
\end{figure}

\subsection{DDPG-H and TD3-H}
\begin{figure}[t]
\centering
\includegraphics[width=65mm,height=20mm]{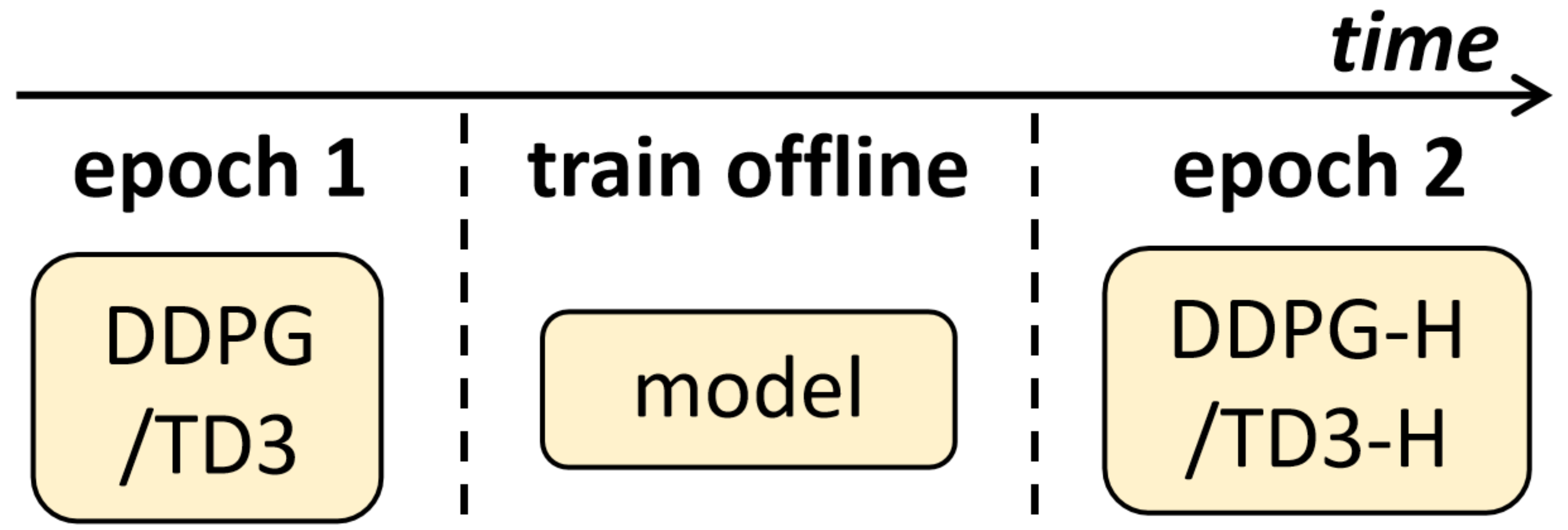}
\caption{\small{Model training timeline. We first train the initial DDPG and TD3. Then, the collected temporary dataset is used to train an avoidance state recognition model offline. Finally, the recognized avoidance state is fed into the initial algorithm (DDPG, TD3) for a new epoch of training (DDPG-H, TD3-H).}}
\label{fig5}
\end{figure}

The UAV will be rewarded when it approaches the secondary target point. When there is no obstacle in a certain range around the UAV, the UAV is considered to be out of the avoidance state, and the initial non-sparse reward function (\emph{i.e.}, Eq. \ref{eq7}) is reset. This idea is simple and feasible in a 2D environment. However, in a 3D environment, we tend to face more complex situations. Fig. \ref{fig4} shows the policy we want the UAV to choose when it faces different 3D obstacles. In short, in some complex situations, e.g., when the obstacle is too high and has difficulty for policy exploration in a short time, we set the UAV in an avoidance state. However, when exploring navigation policies, it is difficult for UAVs to recognize the avoidance state. Therefore, we introduce human experts to assist in the avoidance state recognition. Fig. \ref{fig5} depicts the timeline for different models during training.

First, deep reinforcement learning training is performed on the relevant collected data during training. When humans think that the UAV is in an obstacle avoidance state, a label $1$ is set for the temporarily collected data. When humans think the UAV is in a normal state, a label $0$ is set for the temporarily collected data. We collect several groups of scene features and the human labels (2000 groups are taken in this experiment, and one group contains 5 steps) to form a temporary dataset. We train a fully connected neural network model offline on this temporary dataset. When training for new reinforcement learning, the trained models with human labels can help the UAV to identify its state at each step. By changing the reward function in the \emph{avoidance state}, the UAV is guided to avoid obstacles more efficiently.

\section{Experimental Analysis}

\subsection{Experimental Settings}

We build a large-scale complex 3D scene through UE4. As shown in Fig. \ref{fig6}, it contains the 3D urban, rural, and forest scenes. At the beginning of each round of training, the position of the UAV is initialized in the scene center and a target point is randomly generated. The initial reward parameters are set as: $\sigma=8, \eta=-12, \psi=15, \xi=0.4, \delta=0.02, \lambda=100, \alpha_1 =0.1, \alpha_2 =0.2, \beta_1 =10, \beta_2 =20$.

In this paper, DDPG-H and TD3-H are trained with a total of 1500 episodes, in which each episode has a maximal number of 600-time steps. All network parameters are updated with 256 batch size, and Adam optimizer is used with the learning rate of 0.001 for the Actor and 0.0001 for the Critic network, respectively. The discounted factor is $\gamma=0.98$, and the soft target update rate is $\epsilon=0.02$. In addition, an exploration noise with a uniform distribution $U~(-0.25, 0.25)$ is utilized to explore the state and action spaces.

In the experimental part, we will compare the effects of different factors on the model based on different sensor configurations, as well as the effect of the human-in-the-loop mechanism. The specific baselines are defined as follows.
\begin{enumerate}
\item DDPG (distance sensor): Firstly, we only use distance sensors as input to the UAV. A total of 12 distance sensors are placed with yaw angles of [$-90^{\circ}$, $90^{\circ}$] and pitch angles of [$-90^{\circ}$, $60^{\circ}$] with an interval of $30^{\circ}$. We take the data collected by 12 distance sensors, the internal state of the UAV, and the relationship with the target point as the final state: $s_1 = [d_1...d_{12}, \phi, \zeta]$.
\begin{figure}[!t]
\centering
\includegraphics[width=85mm,height=25mm]{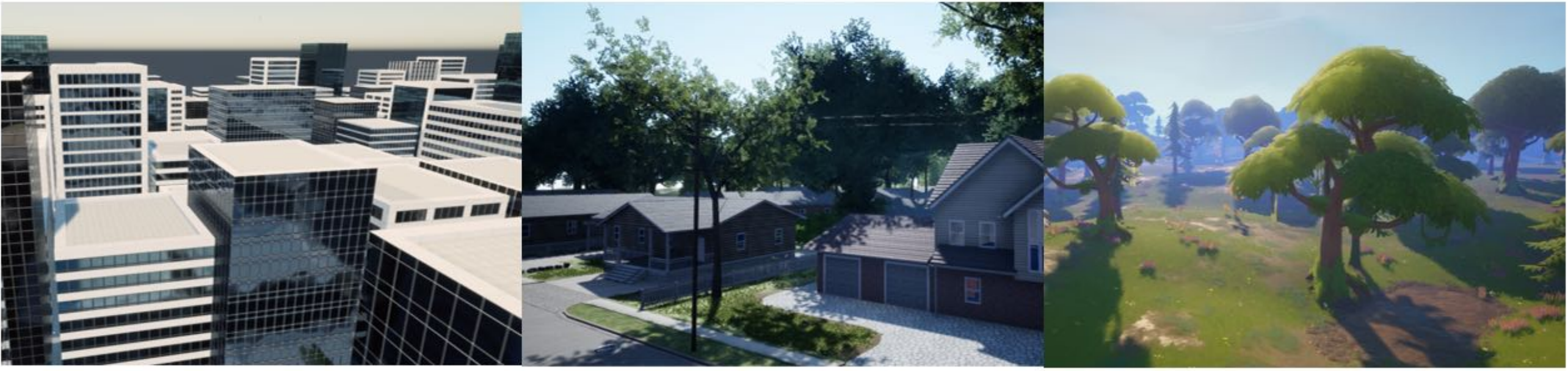}
\caption{3D urban, rural, and forest scenes in UE 4.}
\label{fig6}
\end{figure}
\begin{figure*}
\centering
\includegraphics[width=170mm,height=40mm]{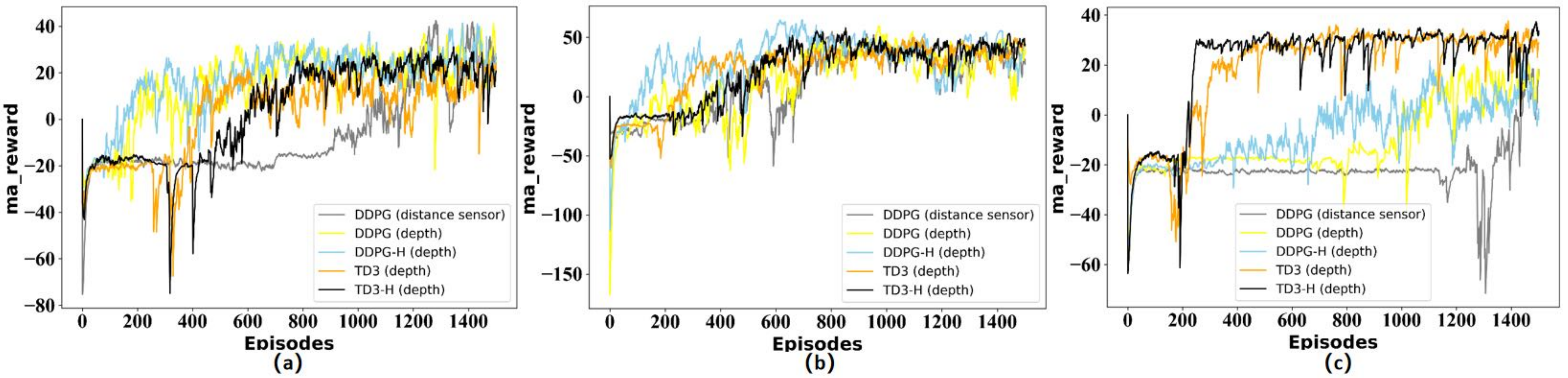}
\caption{Curves of exponential moving average rewards in 3D urban (a), rural (b), and forest (c) scenes.}
\label{fig7}
\end{figure*}
\begin{figure*}[!t]
\centering
\includegraphics[width=160mm,height=33mm]{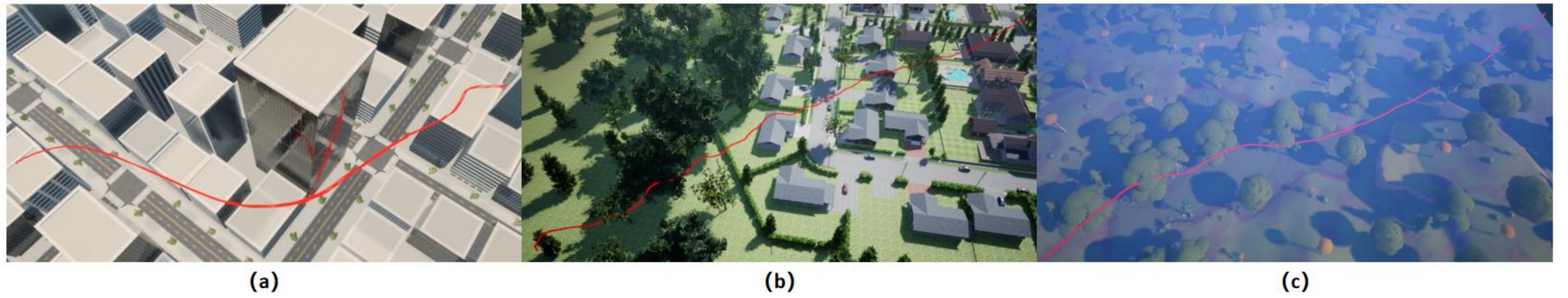}
\caption{The navigated UAV trajectories in 3D urban (a), rural (b), and forest (c) scenes.}
\label{fig8}
\end{figure*}

 \item DDPG (depth) and TD3 (depth): Secondly, the depth map captured by the depth camera is used as input, and the depth map is encoded to a latent feature vector $[c_1...c_{40}]$ by the pre-trained CNN model. In addition, we add vertical distance sensor data $d_1$ into the depth image feature. At this time, the state of the UAV can be characterized as $s_2 = [d_1, c_1...c_{40}, \phi, \zeta]$. 

 \item DDPG-H and TD3-H: Finally, based on the depth map as input, we add a human-in-the-loop model. A marker $f$ is added to the state of the UAV to represent the avoidance state. At this time, the state can be characterized as $s_2 = [d_1, c_1...c_{40}, \phi, \zeta, f]$.
\end{enumerate}
\subsection{Experimental Results}
The curves of exponential moving average reward values in training three baselines are shown in Fig. \ref{fig7}. In Fig. \ref{fig7}, the $x$-axis is the number of episodes and the $y$-axis denotes the exponential moving average reward (ma-reward). 

As shown in Fig. \ref{fig7}(a), the ma-reward value obtained from the urban environment using the DDPG-H reaches the reward value of 45 faster than others at about the 200th episode and then converges around the reward of 45 at about the 600th. In addition, DDPG-H and TD3-H can obtain a larger reward in the training process. In addition, we test three baselines on the success rate (SR.) and weighted average step size (weighted ASS) metrics, respectively. These two indicators are the average value or weighted average value obtained after 100 episodes. The reason we use the weighted average step size is that the navigation distance is not exactly the same for each test. As shown in Fig. \ref{fig7}(a), the exponential reward value using the DDPG-H (depth) reaches the reward value of 25 faster than DDPG (depth) at about the 200th episode and then converges around the reward of 40 at about the 700th. The exponential reward value using the TD3-H (depth) reaches the reward value of 25 faster than TD3 (depth) at about the 300th episode and converges around the reward of 30 at about the 700th episode. Table \ref{tab1} shows the results on the success rate and weighted average step size of three baselines. The difference in efficiency among the three baselines can be visually seen by the weighted average step size. TD3-H (depth) shows better performance on the SR. and DDPG-H (depth) shows better performance on the weighted ASS. Besides, we show the path planning results of the UAV from an aerial view of the urban scene. As shown in Fig. \ref{fig8}(a), the UAV can successfully avoid different obstacles in 3D urban space while approaching the target point.

In addition to the urban scene, we chose two other scenes and train the model separately. One is the rural scene, officially provided by Airsim, which consists of central buildings and peripheral woods. Because the buildings in this environment are generally low, we lower the height of the random target points in training. As shown in Fig. \ref{fig7}(b), the exponential reward value using the DDPG-H (depth) and DDPG (depth) reaches the reward value of 20 at about the 300th episode and then converges around the reward of 25 at about the 700th. While the exponential reward value using the DDPG (distance sensor) reaches the reward value of 20 until about the 1200th and failed to converge. The exponential reward value which using the TD3 (depth) converges around the reward of 10 at about the 1200th episode, and which using the TD3-H (depth) converges around the reward of 20 at about the 800th episode. The test results are shown in Table \ref{tab2}, and the trajectory of the UAV navigation is shown in Fig. \ref{fig8}(b).

Another environment is the forest scene. Similarly, due to the low height of the trees, we reduce the height of random target points. As shown in Fig. \ref{fig7}(c), the exponential reward value using the DDPG-H reaches the reward value of 5 faster than DDPG at about the 800th episode and then converges at about the 1250th episode. The exponential reward value using the TD3 reaches the reward value of 25 faster than TD3 at about the 300th episode and then converges. The test results are shown in Table \ref{tab3}, and the UAV trajectory is shown in Fig. \ref{fig8}(c). In summary, the test results in three scenarios verify that the model can be enhanced by combining the human-in-the-loop mechanism.

\begin{table}[!t]
  \centering
  \caption{The results of success rate (SR.) and weighted average step size (Weighted-ASS) for different baselines in an urban environment with 100 trials.}
  \label{tab1}
  \begin{tabular}{c|c|c}
\toprule[0.8pt]
    Model  & SR.$\uparrow$ & Weighted-ASS$\downarrow$ \\ \hline
    DDPG (distance sensor) \cite{22}  &  0.59 & 412.59 \\ 
    DDPG (depth) \cite{22}     &  0.73 & 264.36 \\ 
    TD3 (depth) \cite{23}        &  0.85 & 287.52        \\
     \hline
    DDPG-H (depth)   &  0.92 & \textbf{179.77} \\
    TD3-H (depth)         &  \textbf{0.94} &  238.44 \\
    \hline
  \end{tabular}
\end{table}

\begin{table}[!t]
  \centering
  \caption{The results of success rate (SR.) and weighted average step size (Weighted-ASS) for different baselines in a rural environment with 100 trials.}
  \label{tab2}
  \begin{tabular}{c|c|c}
\toprule[0.8pt]
    Model  & SR.$\uparrow$ & Weighted-ASS$\downarrow$\\ \hline
    DDPG (distance sensor) \cite{22}  &  0.39 & 317.33\\ 
    DDPG (depth) \cite{22}  &  0.57 & 201.90  \\  
    TD3 (depth) \cite{23}    &  0.51 &  239.31  \\
    \hline
    DDPG-H (depth)       &  \textbf{0.65} & \textbf{177.38}\\
    TD3-H (depth)     &  0.62 &  218.32  \\   
    \hline
  \end{tabular}
\end{table}

\begin{table}[!t]
  \centering
  \caption{The results of success rate (SR.) and weighted average step size (Weighted-ASS) for different baselines in a forest environment with 100 trials.}
  \label{tab3}
  \begin{tabular}{c|c|c}
\toprule[0.8pt]
    Model  & SR.$\uparrow$ & Weighted-ASS$\downarrow$\\ \hline
    DDPG (distance sensor) \cite{22}  & 0.57  & 317.33 \\ 
    DDPG (depth) \cite{22}   &  0.75 & 205.43 \\ 
    TD3 (depth) \cite{23}           &  0.80 &  236.44      \\
    \hline
    DDPG-H (depth)   &  0.81 & 224.58\\
    TD3-H (depth)    &  \textbf{0.85} & \textbf{185.25}      \\
    \hline
  \end{tabular}
\end{table}

From the evaluation, the results show that TD3-H performs best in the forest scene, DDPG-H performs best in the rural scene, and these two methods have their own merits in the urban scene. We think the possible reason for this result is that in rural scenes, there are some large trees. When the UAV passes through tree leaves, it may cause collision problems for the navigation strategy of the UAV. Therefore, scene construction and collision detection in the training process is also a key issue to be concentrated. 

\section{Conclusion}

In this paper, we develop a human-in-the-loop reinforcement learning framework for UAV navigation in large-scale 3D complex environments. By introducing human expert knowledge to assist the recognition of avoidance state in navigation, the success rate and the navigation efficiency are improved significantly. We evaluate the performance on urban, rural, and forest scenes, and the results show that TD3-H and DDPG-H perform better than others. In the future, we will improve the human-in-the-loop method and explore the human-in-the-loop mechanism in group UAV navigation.

\bibliographystyle{unsrt}\footnotesize
\bibliography{ref}

\begin{thebibliography}{10}

\bibitem{1}
Farhan Mohammed, Ahmed Idries, Nader Mohamed, Jameela Al-Jaroodi, and Imad
  Jawhar.
\newblock Uavs for smart cities: {O}pportunities and challenges.
\newblock In {\em International Conference on Unmanned Aircraft Systems}, pages
  267--273, 2014.

\bibitem{2}
Jin Liu et~al.
\newblock Research on low-altitude {UAV} aerial photography target detection.
\newblock In {\em International Conference on Computer Network, Electronic and
  Automation}, pages 369--372, 2022.

\bibitem{3}
Stavroula Charalampidou, Eleftherios Lygouras, Ioannis Dokas, Antonios
  Gasteratos, and Aikaterini Zacharopoulou.
\newblock A sociotechnical approach to {UAV} safety for search and rescue
  missions.
\newblock In {\em International Conference on Unmanned Aircraft Systems}, pages
  1416--1424, 2020.

\bibitem{4}
Pooja Agrawal, Ashwini Ratnoo, and Debasish Ghose.
\newblock Inverse optical flow based guidance for {UAV} navigation through
  urban canyons.
\newblock {\em Aerospace Science and Technology}, 68:163--178, 2017.

\bibitem{5}
Xin-Zhong Peng, Huei-Yung Lin, and Jyun-Min Dai.
\newblock Path planning and obstacle avoidance for vision guided quadrotor
  {UAV} navigation.
\newblock In {\em IEEE International Conference on Control and Automation},
  pages 984--989, 2016.

\bibitem{6}
Yintao Zhang, Youmin Zhang, Zhixiang Liu, Ziquan Yu, and Yaohong Qu.
\newblock Line-of-sight path following control on {UAV} with sideslip
  estimation and compensation.
\newblock In {\em Chinese Control Conference}, pages 4711--4716, 2018.

\bibitem{8}
A~Moura, J~Antunes, Andr{\'e} Dias, Alfredo Martins, and Jos{\'e} Almeida.
\newblock Graph-slam approach for indoor {UAV} localization in warehouse
  logistics applications.
\newblock In {\em IEEE International Conference on Autonomous Robot Systems and
  Competitions}, pages 4--11, 2021.

\bibitem{9}
Pengtao Shao, Fan Mo, Yaqian Chen, Ning Ding, and Rui Huang.
\newblock Monocular object slam using quadrics and landmark reference map for
  outdoor {UAV} applications.
\newblock In {\em IEEE International Conference on Real-time Computing and
  Robotics}, pages 1195--1201, 2021.

\bibitem{10}
Mirco Theile, Harald Bayerlein, Richard Nai, David Gesbert, and Marco Caccamo.
\newblock {UAV} coverage path planning under varying power constraints using
  deep reinforcement learning.
\newblock In {\em IEEE/RSJ International Conference on Intelligent Robots and
  Systems}, pages 1444--1449, 2020.

\bibitem{11}
Yang Gao, Yuankai Li, and Ziqi Guo.
\newblock A {Q}-learning based {UAV} path planning method with awareness of
  risk avoidance.
\newblock In {\em China Automation Congress}, pages 669--673, 2021.

\bibitem{12}
Tianze Zhang, Xin Huo, Songlin Chen, Baoqing Yang, and Guojiang Zhang.
\newblock Hybrid path planning of a quadrotor {UAV} based on {Q}-learning
  algorithm.
\newblock In {\em Chinese control conference}, pages 5415--5419, 2018.

\bibitem{13}
Sihem Ouahouah, Miloud Bagaa, Jonathan Prados-Garzon, and Tarik Taleb.
\newblock Deep-reinforcement-learning-based collision avoidance in uav
  environment.
\newblock {\em IEEE Internet of Things Journal}, 9(6):4015--4030, 2021.

\bibitem{14}
Chao Yan, Xiaojia Xiang, and Chang Wang.
\newblock Towards real-time path planning through deep reinforcement learning
  for a {UAV} in dynamic environments.
\newblock {\em Journal of Intelligent \& Robotic Systems}, 98(2):297--309,
  2020.

\bibitem{15}
Guan-Ting Tu and Jih-Gau Juang.
\newblock Path planning and obstacle avoidance based on reinforcement learning
  for {UAV} application.
\newblock In {\em International Conference on System Science and Engineering},
  pages 352--355, 2021.

\bibitem{16}
Sanghyun Kim, Jongmin Park, Jae-Kwan Yun, and Jiwon Seo.
\newblock Motion planning by reinforcement learning for an unmanned aerial
  vehicle in virtual open space with static obstacles.
\newblock In {\em International Conference on Control, Automation and Systems},
  pages 784--787, 2020.

\bibitem{17}
Bilal Kabas.
\newblock Autonomous uav navigation via deep reinforcement learning using
  {PPO}.
\newblock In {\em Signal Processing and Communications Applications
  Conference}, pages 1--4, 2022.

\bibitem{18}
Fawad~Salam Khan, Mohd Norzali~Haji Mohd, Raja~Masood Larik, Muhammad~Danial
  Khan, Muhammad~Inam Abbasi, and Susama Bagchi.
\newblock A smart flight controller based on reinforcement learning for
  unmanned aerial vehicle ({UAV}).
\newblock In {\em IEEE International Conference on Signal and Image Processing
  Applications}, pages 203--208, 2021.

\bibitem{19}
Hao Xie, Dingcheng Yang, Lin Xiao, and Jiangbin Lyu.
\newblock Connectivity-aware {3D} {UAV} path design with deep reinforcement
  learning.
\newblock {\em IEEE Transactions on Vehicular Technology}, 70(12):13022--13034,
  2021.

\bibitem{20}
Yibing Li et~al.
\newblock A uav path planning method based on deep reinforcement learning.
\newblock In {\em IEEE USNC-CNC-URSI North American Radio Science Meeting
  (Joint with AP-S Symposium)}, pages 93--94, 2020.

\bibitem{21}
Chao Wang, Jian Wang, Yuan Shen, and Xudong Zhang.
\newblock Autonomous navigation of uavs in large-scale complex environments: A
  deep reinforcement learning approach.
\newblock {\em IEEE Transactions on Vehicular Technology}, 68(3):2124--2136,
  2019.

\bibitem{22}
Timothy~P Lillicrap, Jonathan~J Hunt, Alexander Pritzel, Nicolas Heess, Tom
  Erez, Yuval Tassa, David Silver, and Daan Wierstra.
\newblock Continuous control with deep reinforcement learning.
\newblock {\em arXiv preprint arXiv:1509.02971}, 2015.

\bibitem{23}
Fujimoto et~al.
\newblock Addressing function approximation error in actor-critic methods.
\newblock In {\em International Conference on Machine Learning}, pages
  1587--1596, 2018.

\bibitem{24}
Shital Shah, Debadeepta Dey, Chris Lovett, and Ashish Kapoor.
\newblock Airsim: High-fidelity visual and physical simulation for autonomous
  vehicles.
\newblock In {\em Field and Service Robotics}, pages 621--635, 2018.

\bibitem{25}
Weichao Qiu and Alan Yuille.
\newblock Unrealcv: Connecting computer vision to unreal engine.
\newblock In {\em European Conference on Computer Vision}, pages 909--916,
  2016.

\bibitem{26}
Rishabh Agarwal, Chen Liang, Dale Schuurmans, and Mohammad Norouzi.
\newblock Learning to generalize from sparse and underspecified rewards.
\newblock In {\em International Conference on Machine Learning}, pages
  130--140, 2019.

\bibitem{27}
Chao Wang, Jian Wang, Jingjing Wang, and Xudong Zhang.
\newblock Deep-reinforcement-learning-based autonomous {UAV} navigation with
  sparse rewards.
\newblock {\em IEEE Internet of Things Journal}, 7(7):6180--6190, 2020.

\bibitem{28}
Chiya Zhang, Shiyuan Liang, Chunlong He, and Kezhi Wang.
\newblock Multi-uav trajectory design and power control based on deep
  reinforcement learning.
\newblock {\em Journal of Communications and Information Networks},
  7(2):192--201, 2022.

\bibitem{29}
B~Ravi Kiran, Ibrahim Sobh, Victor Talpaert, Patrick Mannion, Ahmad~A
  Al~Sallab, Senthil Yogamani, and Patrick P{\'e}rez.
\newblock Deep reinforcement learning for autonomous driving: A survey.
\newblock {\em IEEE Transactions on Intelligent Transportation Systems},
  23(6):4909--4926, 2022.

\end{thebibliography}

\end{document}